\documentclass[runningheads]{llncs}
\usepackage[T1]{fontenc}
\usepackage{tabularx}
\usepackage{graphicx}
\usepackage{hyperref}
\usepackage{color}

\begin{document}
\title{Fine-Tuning BERTs for Definition Extraction from Mathematical Text}
\titlerunning{Definition Extraction from Mathematical Text}
%
\author{Lucy Horowitz \and
Ryan Hathaway}
\authorrunning{L. Horowitz and R. Hathaway}
%
\institute{
University of Chicago, Chicago IL 60637, USA\\
\email{\{lucyhorowitz,ryanhathaway\}@uchicago.edu}}
\maketitle              
\begin{abstract}
In this paper, we fine-tuned three pre-trained BERT models on the task of ``definition extraction'' from mathematical English written in \LaTeX. This is presented as a binary classification problem, where either a sentence contains a definition of a mathematical term or it does not. We used two original data sets, ``Chicago'' and ``TAC'', to fine-tune and test these models. We also tested on WFMALL, a dataset presented by Vanetik and Litvak~\cite{vanetikDefinitionExtractionGeneric2021} and compared the performance of our models to theirs. We found that a high-performance Sentence-BERT transformer model performed best based on overall accuracy, recall, and precision metrics, achieving comparable results to the earlier models with less computational effort. A repository of our fine-tuned models and data preprocessing can be found \color{blue}\href{https://drive.google.com/drive/folders/1A1hPhwvGQKIi1vHCAYRrmyq_oZmXBurV?usp=drive_link}{here}\color{black}.

\end{abstract}
\section{Introduction}
Definitions are foundational to mathematics—they outline almost every concept a mathematician might care about. Most modern mathematics is written in \LaTeX, a typesetting language which allows the writer to neatly display equations and format PDF documents. Sometimes in the course of this formatting, definitions are ``called out'' by an environment that tells the reader: ``I'm defining a new concept, so pay attention!'' Sometimes, however, a definition is undifferentiated and simply embedded in the middle of a paragraph of text.
 
As the mathematics literature grows, it becomes harder and harder for any individual mathematician to stay on top of recent research. In particular, it can be difficult to figure out whether something one is working on has already been proved, or whether a new concept one is thinking of defining has already been used and explored. A researcher's time is valuable, but one often spends much of it searching fruitlessly through the literature or even rediscovering and re-proving known results that are just hard to find~\cite{wolchoverNeutrinosLeadUnexpected2019}. The discipline and the productivity of researchers would benefit greatly from a searchable library of concepts and results, but creating such a library manually is out of the question given the sheer size of the literature. Therefore an automated solution must be devised. We propose to complete one step in automating the process: using machine learning to extract natural language definitions of math concepts from \LaTeX{} code. 

We use pre-trained BERT models fine-tuned on two datasets of definitional and non-defnitional sentences about mathematics to perform this task as a binary classification problem and compare their respective performances. This required us to make some decisions about what should count as a definition. As explained in section \ref{chicago}, sometimes the full definition of a concept contains multiple sentences. They should not all be annotated as definitional because some of them do not \textit{actually} make up a definition of that concept. Kruse et al.~\cite{kruseDefinitionExtractionMathematical2021} refer to certain definitions requiring anaphora resolution (roughly the substitution of a pronoun for the thing to which it refers) to be full definitions of the concept. They do not consider these definitions, while we do. This is because in the future, we hope to improve a system called MathGloss~\cite{horowitzMathGlossBuildingMathematical2023} that will provide pointers to definitions in mathematical texts—if a definition requiring anaphora resolution can be automatically extracted, the user or reader will find the appropriate context when directed to the location in text where it occurs.

\section{Related Work}
The general problem of definition extraction is a longstanding one~\cite{singhDSCIITISMSemEval20202020}, but only recently have we seen projects aimed specifically at extracting definitions from mathematical text. One is Vanetik and Litvak's work on definition extraction, which also involved the creation of a dataset of mathematical definitions, called WFMALL~\cite{vanetikDefinitionExtractionGeneric2021}. The original text of WFMALL comes from the Wolfram MathWorld\footnote{\url{https://mathworld.wolfram.com}} articles that are not about the Wolfram programming language. It contains 6,140 sentences, of which 1,934 are annotated as definitional for a ``definition density'' of 31.5\%. The other datasets they test on have similar proportions of definitions, but all less than 40\%. 

Vanetik and Litvak created several neural network models to perform definition extraction from mathematical text. Their models perform very well on WFMALL and reasonably well on other common datasets for definitions. They found that working within one domain (e.g. in mathematical language) oversampling both universally improved the performance of their models. Their best-performing models achieved an accuracy of 0.922, and taken together in an ensemble, all of the models collectively achieved an accuracy of 0.942. They did not report precision or recall. We perform some experiments with WFMALL and take inspiration from them for other experiments. 

Kruse et al.~\cite{kruseDefinitionExtractionMathematical2021} performed definition extraction from a corpus of mathematical text that specifically focuses on the field of graph theory in both German and English. They use different techniques for each language; we focus on their English experiments here. They trained their neural network on an earlier version of the WFMALL dataset~\cite{vanetikAutomatedDiscoveryMathematical2020} which contains fewer sentences and tested on a manually collected set of 100 definitions and 100 non-definition sentences. On this test, they achieved a precision of 0.7054 and a recall of 0.91. They do not report accuracy, which makes comparison between the original tests of this dataset somewhat difficult.

\section{Datasets}
We present two annotated datasets of definitions: ``Chicago'' and ``TAC.'' Both are written with \LaTeX{} commands to display equations and mathematical notation. The next two subsections provide descriptions of these two datasets. 

\subsection{Chicago}\label{chicago}
The Chicago dataset consists of definitions of mathematical terms encountered by the first author over the course of her undergraduate degree in mathematics at the University of Chicago. There were 702 different markdown files each corresponding to a mathematical concept. The definitions of these concepts collectively contained 1,599 sentences. They included sentencs explaining and elaborating ideas as well as sentences that serve to set up the terminology or notation used in a definition. We annotated 814 definitional sentences, for a ``definition density'' of 50.1\%. There are more annotated definitional sentences than ``concepts,'' if we count concepts by number of files, because some files contained definitions of multiple concepts: for example, in the file that defines ``group homomorphism,'' there is a sentence that says ``A group isomorphism is a homomorphism that is also bijective.'' 

\subsection{TAC}
The TAC dataset consists of 755 abstracts from the journal \textit{Theory and Applications of Categories}\footnote{\url{http://www.tac.mta.ca/tac/}} up to 2020 and in total contains 3,068 sentences. Often the abstracts included mathematical notation that was not contained in any of the math mode delimiters, so we manually added them in where they were missing. We annotated 231 definitional sentences for a definition density of 7.53\%. It is not unreasonable to have so few definitions in this dataset as it is more common in research mathematics to introduce a new result or theorem than a new concept. The authors are not domain experts in category theory, so it is possible that we annotate theorems that look like definitions (but are not) as definitional. To take an example from~\cite{kruseDefinitionExtractionMathematical2021}, the statement ``A connected graph with $n$ vertices and $n-1$ edges is a tree'' may look like a definition of ``trees,'' but in fact it is a statement about them.

\subsection{Cleaning the text}
The original sources of the Chicago and TAC datasets were in \LaTeX{}, which includes delimiters \verb|$-$| as well as \verb|$$-$$|, \verb|\[-\]|, and \verb|\(-\)| to indicate ``math mode'' and between which an author can use many commands to render symbols. The raw \LaTeX{} also includes environments like \verb|\begin{align} -\end{align}|, formatting indicators for boldface and italics, chapter or section headers, and sometimes citations.

The \LaTeX{} commands for formatting are not relevant to the content or form of a definition\footnote{Although in the original text of the Chicago notes, words being defined were always bolded in Markdown syntax. It is not desirable to have this formatting in the dataset itself because our goal is generalization, and not every author bolds their definienda or wraps their definitions in environments.} but the commands in math mode could very well be: formulas often do much of the heavy lifting in conveying the meaning of a mathematical statement. Thus when cleaning the data, we removed \LaTeX{} commands for formatting but left math mode commands as they were. 

However, it was necessary to make some changes to the math mode commands because of the problem of ``sentencization.'' The text from the corpora was not already split into sentences, so we had to create a reliable way to do this. It would have been too much to do automatically, so we developed a pipeline component spaCy~\cite{Honnibal_spaCy_Industrial-strength_Natural_2020} that was able to perform sentencization on reliably on \LaTeX{} with formatting commands removed. Initially, \LaTeX{} code often caused spaCy to make sentencization errors: the appearance of non-word commands in a string would result in sentences both blending together and being split apart at inappropriate locations. 

There were a few key changes made to the \LaTeX{} code in the corpora. First was to standardize the math mode delimiters. We changed each of the four types listed above to single dollar signs, and also ensured that any punctuation that was part of the sentence was placed \textit{outside} the final delimiter. The second was to change any appearances of an exclamation mark to a placeholder string—we used ``clik.'' Mathematical writing usually takes on a formal tone and does not often use exclamation points at the end of sentences to indicate excitement. The final important change was to pad every hyphen with a space. This enabled spaCy to correctly tokenize terms like \verb|$G$-space|, which occurred often in the two corpora.

After cleaning the text as above, we created a new pipeline component for spaCy called \verb|detextor|. The \verb|detextor| component essentially tells spaCy to treat everything in between dollar sign delimiters as a single token. It is implemented after the \verb|tagger| component of spaCy's default pipeline, which assigns parts of speech to tokens. More on cleaning and the \verb|detextor| component can be found in~\cite{horowitzMathGlossBuildingMathematical2023}. 
    
Together, cleaning the text and implementing \verb|detextor| made spaCy almost perfect at splitting \LaTeX{} into sentences. When there were errors in sentencization, (not more than 30 out of thousands) we were able to correct them manually during the annotation process.

Each of the two corpora required a little bit of ``special treatment'' due to its own idiosyncrasies. Chicago was written as several hundred individual Markdown files, one for each concept defined. It contained links between files that would point a reader to another concept if it was referenced. As an example of a link between files, if one wanted to link to the definition of ``group'' one would write \verb|[[group]]|. If, on the other hand, one needed to use the plural ``groups'' to make the sentence correct or grammatical, one would write \verb=[[group|groups]]= to make the rendered Markdown file read simply a hyperlinked ``groups.'' As part of the cleaning process for Chicago, we also removed Markdown formatting and any links between files, preserving grammaticality (i.e. leaving behind only the part of a link after a pipe, if it exists). Some of the abstracts in TAC had mathematical notation but no math mode delimiters. The first author manually added these delimiters (single dollar signs) where they were missing. We found no difference in performance between the performance of the models when trained on the text including the dollar sign delimiters and when trained on the text \textit{not} including the dollar sign delimiters, so we use the former because it is closer to the original source of the text. 

\section{Identifying Pre-trained Models}

We experimented with three pre-trained BERT models available for use in the hugging-face repositories. Ideally, models trained for math-related NLP tasks would perform optimally in our classification task. For a point of comparison, we included a model not trained on math-related tasks. We implemented three models for fine-tuning.

\subsection{MathBERT}

Shen et al.'s MathBERT~\cite{shenMathBERTPretrainedLanguage2023} is a pre-trained transformer model that was trained on a variety of mathematical text ranging in level from pre-K to graduate mathematics. They tested it by fine-tuning for a few tasks relevant in mathematical education: for example, grading open-ended answers to math problems. The developers explain that it is appropriate for general NLP tasks of mathematical content, so it is not optimized for binary definition classification. We identified this model as a potential high performer due to its general math training which could be leveraged for our more specific task.

Shen et al. also created a second model called MathBERT-custom. While MathBERT was pre-trained with the original vocabulary from BASE BERT, MathBERT-custom was pre-trained with a custom vocabulary that included terms that while common in mathematical language, would not often appear in everyday text. They find that after fine-tuning, MathBERT-custom performs worse than the original MathBERT, which aligns with our results (see Tables \ref{table1} and \ref{table2}). 

\subsection{Sentence-BERT}

Reimers and Gurevych's Sentence-BERT~\cite{reimersSentenceBERTSentenceEmbeddings2019}, also called SBERT, is a state-of-the-art pre-trained network for semantic similarity tasks. Although the authors say that transfer learning is not the purpose of SentenceBERT, we find fine-tuning it outperforms fine-tuning BASE BERT as performed by Vanetik and Litvak in \cite{vanetikDefinitionExtractionGeneric2021}, which was the proposed alternative in \cite{reimersSentenceBERTSentenceEmbeddings2019}. Moreover, the authors of \cite{reimersSentenceBERTSentenceEmbeddings2019} found that their model outperforms the state of the art on Semantic Textual Similarity tasks when fine-tuned for them.

\section{Methods and Experiments}
Each model supported binary classification. We fine-tuned each using our two datasets in a standardized arrangement implementing a binary cross entropy loss function. Additionally, we pooled examples from both datasets into a combined dataset to record how the variety of definition styles would affect the performance. In this way, we developed a total of three training datasets, with 20\% of each set of sentences saved for validation. 

For each of the three training sets, Chicago, TAC, and the combined set, the annotated sentences were padded then truncated to a max length of 512 characters to standardize lengths and leverage positional encoding. Very few definitions were longer than 512 characters. We tokenized all inputs and validation sentences using the pre-made tokenizer for each respective pre-trained model, applied an attention mask, and initialized the model for binary classification. 

Each dataset was treated with the same preprocessing hyper-parameters. In particular, for each training, we split the dataset into batch sizes of 10 examples and fine-tuned for 3 epochs with a standard cross-entropy loss function. To save on computational resources, we trained all models and validated all results using a virtual GPU runtime environment through Google Colab. By leveraging the NVIDIA A100 Tensor Core GPU in the virtual runtime environment no computation time exceeded 6 minutes and the average length for a fine-tuning runtime was 3 minutes. We recorded standard performance metrics for each model on each dataset, giving precedence to the recall metric as we are most concerned about not missing true positives.  

We ran several experiments, varying the model, the dataset, whether or not we oversampled the dataset, and whether or not we included dollar sign delimiters in the dataset. We also tested our models on the WFMALL dataset. Due to time constraints, we were unable to perform all of the permutations. The results of early experiments informed which experiments we would run next. 

With respect to computing resources, all models took a negligible time of no more than three minutes to fine-tune using GPU acceleration on Google Colab. Leveraging the pre-trained models, we were able to produce relatively high-performance models for our classification task efficiently.

\section{Results}

The first set of experiments was to fine-tune each of the three BERT models on Chicago, TAC, and then the combined Chicago and TAC dataset with no oversampling. Tables \ref{table1}, \ref{table2}, and \ref{table3} record the performance of our model fine-tuning on each of the three datasets. 

\begin{table}[htbp]
\centering
\caption{Fine-Tuning MathBert}
\label{table1}
\footnotesize
\begin{tabularx}{\columnwidth}
{|X|X|X|X|X|}
\hline
Dataset & Accuracy & Precision& Recall & $F_1$\\
\hline\hline
Chicago & 0.913 & 0.904 & 0.928 & 0.916 \\
\hline
TAC & 0.968	&0.667	&0.417	&0.513 \\
\hline
Combo & 0.952&	0.894&	0.846&0.869 \\
\hline
\end{tabularx}
\label{tab:mbert1_performance}
\
\caption{Fine-Tuning MathBert-custom}
\label{table2}
\footnotesize
\begin{tabularx}{\columnwidth}{|X|X|X|X|X|}
\hline
Dataset & Accuracy & Precision& Recall & $F_1$ \\
\hline\hline
Chicago & 0.903	&0.887&	0.928&	0.907 \\
\hline
TAC & 0.965&	0.600	&0.375&	0.462 \\
\hline
Combo & 0.923	&0.825&	0.774	&0.799 \\
\hline
\end{tabularx}
\label{tab:mbert2_performance}
\
\caption{Fine-Tuning SentenceBERT}
\label{table3}
\footnotesize
\begin{tabularx}{\columnwidth}{|X|X|X|X|X|}
\hline
Dataset & Accuracy & Precision& Recall & $F_1$ \\
\hline\hline
Chicago & 0.933	&0.917&	0.963	&0.939 \\
\hline
TAC & 0.975	&1.000&	0.400&	0.571 \\
\hline
Combo & 0.950	&0.859	&0.893&	0.875 \\
\hline
\end{tabularx}
\label{tab:sbert_performance}
\end{table}

While all the models had very high accuracy, they had much lower recall when trained only on the TAC dataset. Downstream applications of definition extraction would suffer more from false negative classifications than from false positive classifications (which might still contain useful information about mathematical concepts), this is undesirable in a model. It is likely that this is due to the low density of definitions in the TAC dataset (approximately 7\%). 

We tested how the SBERT models performed on the WFMALL dataset, with results in Table \ref{wfmalltest}.  The performance declines somewhat when generalizing, and this may be a result of overfitting. However, the performance is still reasonable, especially when trained on the Combination dataset. Thus these models could still prove very useful for extracting definitions from new sources.

\begin{table}[htbp]
   \centering
   \caption{Testing SBERT on WFMALL}\label{wfmalltest}
   \footnotesize
   \begin{tabularx}{\columnwidth}{|X|X|X|X|X|X|X|}\hline
   Dataset & Accuracy & Precision & Recall & $F_1$ & Support \\
        \hline\hline
Chicago & 0.79 & 0.59 & 0.91 & 0.72 & 208 \\\hline
        TAC & 0.82 & 0.86 & 0.50 & 0.64 & 212 \\\hline
        Combo & 0.87 & 0.79 & 0.76 & 0.77 & 201 \\\hline
  \end{tabularx}
\end{table} 

We also tested the two MathBERT models on WFMALL, but they performed incredibly poorly—they had recall of zero or near-zero, which is certainly unacceptable for any application. In the experiments that follow, we focus on SentenceBERT models due to their better ability to generalize.

Vanetik and Litvak \cite{vanetikDefinitionExtractionGeneric2021} achieved an accuracy of 0.760 when fine-tuning BASE BERT on the WFMALL dataset, and found that oversampling (using the ``minority'' setting from \verb|imbalanced-learn|\footnote{\url{https://imbalanced-learn.org/stable/}}, which randomly boosts the proportion of definitions in the dataset to 50\%) actually lowered this to 0.750. We fine-tuned SentenceBERT using the WFMALL dataset both minority oversampled and in its original form and achieved the results summarized in Table \ref{wfmallover}. 

\begin{table}[htbp]\caption{Fine-Tuning SBERT on WFMALL}\label{wfmallover}
    \centering
    \begin{tabularx}{\textwidth}{|X|X|X|X|X|} 
        \hline
        Dataset & Accuracy & Precision & Recall & $F_1$ \\
        \hline\hline
        WFMALL & 0.910 & 0.838 & 0.882 & 0.860 \\
        \hline
        WFMALLmin & 0.924 & 0.924 & 0.930 & 0.925 \\
        \hline
    \end{tabularx}
\end{table} In this case, oversampling did improve the performance of the fine-tuned model in all metrics. Additionally, the accuracy is higher than the best performing individual models from~\cite{vanetikDefinitionExtractionGeneric2021}, which saw 0.922. Moreover, the recall is high which is beneficial when we do not want to miss definitions of math concepts. 

We tested how this model was able to generalize to the Chicago and TAC datasets on a random sample of 700 sentences from each. The results of these tests are summarized in Tables \ref{wfmallgen1} and \ref{wfmallgen2}.

\begin{table}[htbp]
   \centering
   \caption{Testing BERT Fine-Tuned on WFMALL}
   \footnotesize
   \begin{tabularx}{\columnwidth}{|X|X|X|X|X|X|X|}
        \hline
        Dataset & Accuracy & Precision & Recall & $F_1$ & Support\\
        \hline \hline
        
         Chicago & 0.87 & 0.95 & 0.81 & 0.87 & 358  \\
        \hline 
        TAC & 0.93 & 0.38 & 0.61 & 0.47 & 33  \\
    
        \hline
    \end{tabularx}

\label{wfmallgen1}
\
\caption{Testing SBERT Fine-Tuned on Oversampled WFMALL}
\label{wfmallgen2}
\footnotesize
   \begin{tabularx}{\columnwidth}{|X|X|X|X|X|X|X|}
        \hline
        Dataset & Accuracy & Precision & Recall & $F_1$ & Support\\
        \hline\hline
        
         Chicago & 0.89 & 0.92 & 0.86 & 0.89 & 358  \\
        \hline 
        TAC & 0.93& 0.37 & 0.83 & 0.72 & 31 \\
    
        \hline
    \end{tabularx}
\end{table} 

While the changes in the metrics could indicate that these models are overfit, it is likely that the change in performance on the TAC dataset in particular is simply a result of its drastic class imbalance. In this case, oversampling has less of an effect than it does on Vanetik and Litvak's models. The tests performed here are similar in nature to the ``cross-domain''  tests in~\cite{vanetikDefinitionExtractionGeneric2021}, and we achieved higher accuracy than each of them. However, the cross domain tests in ~\cite{vanetikDefinitionExtractionGeneric2021} measured the performance of models trained on mathematical text on general text and vice versa, while these tests stayed within the mathematical domain. 

We also compare to the tests performed by Kruse and Kliche in~\cite{kruseDefinitionExtractionMathematical2021}. They trained a neural network on an older version of the WFMALL dataset~\cite{vanetikAutomatedDiscoveryMathematical2020} and found that, on a sample of 100 randomly selected definitions and 100 randomly selected non-definitions from a corpus of textbooks and papers in the subfield of graph theory, their models achieved a precision of 0.705 and a recall of 0.910. They did not report the accuracy.
Our models had higher precision on the Chicago dataset and lower precision on the TAC datset. In both cases, they had lower recall. However, on Chicago, the $F_1$ scores of our models were higher. 

We also investigated the effects of oversampling on the TAC and combination datasets. We did not oversample Chicago because more than 50\% of the sentences in that dataset were annotated as definitions, so oversampling would have actually increased the number of non-definitions. We oversampled using the ``minority'' setting as before, and we also oversampled to a definition density of 30\% as a conservative comparison to the datasets tested in~\cite{vanetikDefinitionExtractionGeneric2021}. The results of these tests are in Table \ref{oversampledfirst}.

\begin{table}[htbp]
    \centering
    \caption{Fine-Tuning SBERT on Oversampled Data}\label{oversampledfirst}
    \begin{tabularx}{\columnwidth}{|X|X|X|X|X|X|X|}
        \hline
        Dataset & Accuracy & Precision & Recall & F1 \\\hline\hline
        TAC-30\% & 0.92 & 0.94 & 0.69 & 0.80 \\\hline
        TAC-min& 0.98 & 0.96 & 0.98 & 0.97 \\\hline
        Combo-30\% & 0.94 & 0.85 & 0.91 & 0.88 \\\hline
        Combo-min & 0.98 & 0.97 & 0.98 & 0.97 \\
        \hline
    \end{tabularx}
\end{table} Recall that the definition density of TAC is about 7\% and for Combo it is about 22\%. So oversampling for TAC will duplicate a lot of the data and most likely lead to overfitting. Oversampling Combo with the minority setting will also likely lead to overfitting, but oversampling Combo to 30\% might give more reliable results. We tested these models on random samples of 700 sentences from the WFMALL dataset, with results displayed in Table \ref{oversampledtest}.

\begin{table}[htbp]
    \centering
    \caption{Testing SBERT Fine-Tuned on Oversampled Data}\label{oversampledtest}
    \begin{tabularx}{\columnwidth}{|X|X|X|X|X|X|X|}
        \hline
        Trained on  & Accuracy & Precision & Recall & F1 &Support\\\hline\hline
        TAC-30\% & 0.84 & 0.81 & 0.68 & 0.74 &228\\\hline
        TAC-min & 0.84 & 0.91 & 0.58 & 0.68 &222\\\hline
        Combo-30\% & 0.85 & 0.71 & 0.84 & 0.83 &212 \\\hline
        Combo-min & 0.85 & 0.77 & 0.78 & 0.78&234 \\
        \hline
    \end{tabularx}
\end{table}

We can compare these results to those in Table \ref{wfmalltest} to see if oversampling improves the models' performance. In the case of TAC, where oversampling necessarily leads to a lot of duplicate examples, we saw an overall increase in performance on WFMALL. In the combination case, we actually see a slight decrease in performance on all metrics. 

It is useful to look at examples of false negatives and false positives to gain insight into what might cause the model to make mistakes. In table \ref{falses} we provide a few examples.

\begin{table}[htbp]
    \centering
    \caption{MathBERT and SBERT Data}
    \begin{tabularx}{\textwidth}{|c|c|c|X|c|}
        \hline
        \# & Model & Dataset & Sentence & False \\
        \hline\hline
        1 & MathBERT & TAC & Another type of groupoid morphism, called an actor, is closely related to functors between the categories of groupoid actions. & + \\
        \hline
        2 & MathBERT & TAC & To do this, we approximate (X, t) by a flow on a Stone space (that is, a totally disconnected, compact Hausdorff space). & - \\
        \hline
        3 & MathBERT & Chicago & Homotopy equivalence is an equivalence relation. & + \\
        \hline
        4 & MathBERT & Chicago & The index of G is G:H = |G|/|H|. & - \\
        \hline
        5 & SBERT & TAC & (X, tau) is a quasi spectral space. & + \\
        \hline
        6 & SBERT & TAC & When it does, we call the string distributive. & - \\
        \hline
        7 & SBERT & Chicago & Note: When multiplying measures that are infinite, the product is zero if any factor is zero. & + \\
        \hline
        8 & SBERT & Chicago & That is, it is a p-group that is a subgroup of G and that is not a proper subgroup of any other p-subgroup of G. & - \\
        \hline
        9 & SBERT & WFMALL & The symmetric group is a transitive group. & + \\
        \hline
        10 & SBERT & WFMALL & A graph is intrinsically linked iff it contains one of the seven Petersen graphs. & - \\
        \hline
    \end{tabularx}
    \label{falses}
\end{table} Examples 2 and 8 are similar in that the definition of the term is provided after the phrase ``that is.'' The false positives in Examples 1, 3, 7, and 9 provide important information about math concepts, but do not define them—either they give information about a property of something, or describe its relationship to other concepts.

\section{Future Work}

There are many future avenues to explore in the realm of definition extraction for mathematics. Here we viewed the problem of definition extraction as a binary classification problem-we do not attempt to extract the terms being defined, though other models for definition extraction do.  We would like to continue this work and expand into the problem of term extraction i.e. identifying the definiendum from a sentence that has been tagged as a definition. Collard et al. 
~\cite{collardParmesanMathematicalConcept2023} have made progress in extracting important mathematical concepts from undifferentiated text. 

We plan to test how well these models can generalize to other bodies of text, specifically to open-source mathematics textbooks. This is one step in the process of incorporating this work with the expansion of MathGloss~\cite{horowitzMathGlossBuildingMathematical2023}, a glossary of mathematical terms that links them to the Chicago corpus, the nLab, Wikidata's ontology, formalizations in Lean's MathLib, and translations into other languages provided by MuLiMa.\footnote{\url{https://mulima.xyz}}  In particular, we would like to find definitions in the open-source textbooks, and point users who are looking for the definition of a particular term to its appearance in the textbook. 

\section{Conclusion}
In this work, we tested how well pre-trained BERT models were able to perform definition extraction from mathematical text after fine-tuning on several different datasets. The Sentence-BERT models were the highest performers, with reasonable capability to generalize to new mathematical datasets. While the models perform well overall, there is room for improvement. The results are comparable to or better than others in the literature. Moreover, we were able to achieve these results just by fine-tuning pre-existing models, rather than training entire neural networks. We hope this will enable the construction of tools that will make mathematics more searchable and accessible to everyone.

\begin{credits}
\subsubsection{\ackname}
We would like to thank Daniel Lam for running an excellent course on computational linguistics, and for graciously allowing us to continue this work into the spring. We would also like to thank Valeria de Paiva for her guidance and patience. 

\subsubsection{\discintname}
The authors have no competing interests to declare that are
relevant to the content of this article.
\end{credits}
\bibliographystyle{splncs04}
\bibliography{paper}

\end{document}